\documentclass[10pt,twocolumn,letterpaper]{article}

\usepackage{iccv}
\usepackage{times}
\usepackage{epsfig}
\usepackage{graphicx}
\usepackage{amsmath}
\usepackage{amssymb}

\usepackage{kotex}
\usepackage{kotex-logo}
\usepackage{multicol}
\usepackage{verse}
\usepackage{multirow}
\newcommand*{\affaddr}[1]{#1} 
\newcommand*{\affmark}[1][*]{\textsuperscript{#1}}
\newcommand*{\email}[1]{\texttt{#1}}
\newcommand{\inlineeqnum}{\refstepcounter{equation}~~\mbox{(\theequation)}}

\usepackage[pagebackref=true,breaklinks=true,letterpaper=true,colorlinks,bookmarks=false]{hyperref}

\iccvfinalcopy

\def\iccvPaperID{5638}

\begin{document}

\title{RRNet: Repetition-Reduction Network for Energy Efficient Decoder of Depth Estimation}

\author{
Sangyun Oh\affmark[1,2]\thanks{Authors are equally contributed}, Hye-Jin S. Kim\affmark[3,4]\footnotemark[1], Jongeun Lee\affmark[1,2] and Junmo Kim\affmark[3]\\
\affaddr{\affmark[1]School of Electrical and Computer Engineering, UNIST, Ulsan, Korea}\\
\affaddr{\affmark[2]Neural Processing Research Center, Seoul National University, Seoul, Korea}\\
\affaddr{\affmark[3]Robotics Program, School of Electrical Engineering, KAIST, Daejeon, Korea}\\
\affaddr{\affmark[4]Artificial Intelligence Laboratory, ETRI, Daejeon, Korea}\\
\email{\tt\small syoh@unist.ac.kr,  marisan@etri.re.kr, jlee@unist.ac.kr, junmo.kim@kaist.ac.kr} \\
}

\maketitle
\begin{abstract}
We introduce Repetition-Reduction network (RRNet) for resource-constrained depth estimation, offering significantly improved efficiency in terms of computation, memory and energy consumption. 
The proposed method is based on repetition-reduction (RR) blocks. The RR blocks consist of the set of repeated convolutions and the residual connection layer that take place of the pointwise reduction layer with linear connection to the decoder. 
The RRNet help reduce memory usage and power consumption in the residual connections to the decoder layers. 
RRNet consumes approximately 3.84    times less energy and 3.06 times less memory and is approaximately 2.21 times faster, without increasing the demand on hardware resource relative to the baseline network~\cite{17Godard}, outperforming current state-of-the-art lightweight architectures such as SqueezeNet~\cite{iandola2016squeezenet}, ShuffleNet~\cite{zhang2017shufflenet}, MobileNetv2~\cite{sandler2018mobilenet2} and PyDNet~\cite{Poggi18}.
\end{abstract}

\section{Introduction}
Depth estimation is one of core issues for many computer vision applications, including AR/VR localization, robotics, autonomous vehicles, Drones and smart factories. Depth learning approaches~\cite{garg2016unsupervised, zhou2017unsupervised, zhou2017egomotion, casser2019struct2depth, Mahjourian2018egomotion, 18Chang,17Liang,16zbontar,16ranjan,Hui_2018_CVPR} convincingly outperform those with hand-crafted features~\cite{karsch14pami, ladicky14eccv}. However, these approaches require resource-intensive computations, limiting their use in mobile applications that require a lightweight model and utilize relatively low-end graphics processing units(GPUs).

 The most intuitive method of designing a light-weight model is to use light layers with small-sized kernels by suitabley scaling the number of channel parameters using appropriate sub-sampling. Nonetheless, performance can suffer because the trainable data are expensive and the amount of trainable data are limited. 
 Therefore, prior works have implemented various compensation techniques for trainable data~\cite{iandola2016squeezenet,lin2013network,szegedy2015going,szegedy2016rethinking}.
However, previous approaches to light-weight CNN architectures such as MobileNet~\cite{howard2017mobilenet, sandler2018mobilenet2}, SqueezeNet~\cite{iandola2016squeezenet}, and ShuffleNet~\cite{zhang2017shufflenet}, are generally designed for the classification-typed architecture, which we call encoder-only network, narrowing down when the network goes to output. For \emph{encoder-decoder} architectures (e.g., depth estimation and semantic segmentation), current state-of-the-art lightweight CNN models cannot be applied to the decoder architecuture because the model design and data flow differ from those of the encoder-only network, which offsets the model reduction advantages of the encoder structure. 

In general, depth estimation methods require an encoder-decoder architecture. Comparing to classification or detection architectures, the encoder-decoder architectures entail more computational complexity and memory. In addition, the many feature channels in the encoder can lead to extensive computation in the decoder because of residual connections. 
The main challenge is that although deep structures and overlapping information in both encoder and decoder tend to improve performance, such tightly coupled encoder-decoder networks require significant hardware resources in terms of both computation and memory, which hinders efficient deployment in the mobile environment. 
In this paper we address the problem of how to design a lightweight and high-performance encoder-decoder CNN architectures for depth estimation.
In the proposed architecture, the repetition-reduction(RR) blocks can be repeated in either the horizontal or the vertical directions concatenating them using a spatial pyramidal approach~\cite{16ranjan, Hui_2018_CVPR}.    
RR blocks connect a layer in the encoder to one in the decoder similar to skip connections in the UNet(Ronneberger et al.)~\cite{UNet15}. The pointwise layer in the RR block significantly reduce the feature map size.
Verification of the RRNet using real mobile GPU hardware showed the RRNet outperformed current lightweight models in terms of performance, runtime and hardware usage. Moreover, lightweight architecture reduce training time by increasing the GPU utilities. 

\section{Related Work}
Here, we briefly review some of the major approaches to supervised and unsupervised depth estimation methods and lightweight network architectures. 

\textbf{Supervised Depth estimation}
Most depth estimation methods~\cite{18Chang,17Liang,16ranjan,16zbontar, Hui_2018_CVPR} use supervised approaches, which achieve better performance than unspervised methods. In particular, Ranjan et al.~\cite{16ranjan}, Hui et al.~\cite{Hui_2018_CVPR}  and Chang et al.~\cite{psmnet18Chang} used one or two spatial pyramid network using Spatial Pyramid Pooling(SPP)~\cite{spp14He} and obtained SOTA results.
 SPP has been intensively used in encoder-decoder architecture~\cite{spp16Zhao, psmnet18Chang, deeplab32018chen}. In~\cite{deeplab32018chen,atrous16chen}, the SPP module uses adaptive average pooling to compress features into four scales, followed by a pointwise convolution to reduce feature dimension, and different levels of feature maps are concatenated to form the final SPP feature maps.
 
Those good performance in supervised methods are caused by the ground truth data. However, it is very difficult to prepare a training dataset because a human is required to perform detailed ground-truth depth labeling for various camera view-points in a large dataset. Thus, more attention is now being focused on unsupervised learning, which does not require the manual labeling of the dataset by a human.

\textbf{Unsupervised Depth estimation}
Unsupervised depth learning~\cite{garg2016unsupervised, zhou2017unsupervised, zhou2017egomotion, Mahjourian2018egomotion, casser2019struct2depth} offers the benefits of out requiring a pre-training or annotated ground-truth depths which is surprisingly expensive to obtain because of the expensive depth sensors such as LiDARs, Radars and laser scanners. These sensors also have other limitations. LiDAR has shallow channel which hardly cover the full image resolutions. In the case of active sensors such as Kinect and TOF, they have holes around object boundaries and are sensible to strong visible light as well.

In~\cite{casser2019struct2depth, Mahjourian2018egomotion}, unsupervised learning removes the need for separate supervisory signals (depth or ego-motion ground truth, or multi-view video) and they achieve good performance by introducing camera ego-motion in the learning process.
Godard~\cite{17Godard} achieved the good performance in CVPR 2017 due to generating right image from the left image with left-right consistency without ground truth. Godard uses VGG and ResNet architecture and generated decoder similar to the encoder's VGG and ResNet in the inverse way.
 
\textbf{Light-weight Network Designs}
There have been many attempts to make deep neural network lighter -   
Deep compression method~\cite{Han15Deepcompression}, 
Quantization method~\cite{Zhou16_Dorefa}, 
Low rank approaximation method~\cite{Kim15lowrank}, 
Matrix decomposition method~\cite{Lavin15winograd},
Sparse winograd based CNN method~\cite{Liu18winograd}
and so on. 

On the other hand, various light-weight network~\cite{iandola2016squeezenet,howard2017mobilenet, zhang2017shufflenet} propose architectural approaches such as SqueezeNet~\cite{iandola2016squeezenet}, MobileNet v2~\cite{sandler2018mobilenet2} and ShuffleNet~\cite{zhang2017shufflenet}.
SqueezeNet~\cite{iandola2016squeezenet} compress the model by introducing fire module. Fire module divides the input tensor into convolution layers with different kernel sizes of 3 x 3 and 1 x 1, respectively. After then, it concatenates them as an output tensor to reduce weight parameters. This module can be considered multiple fire modules within a network.  
Depthwise Separable Convolution (DWconv)~\cite{dwconv16chollet} is a layer factorization approach that widely used for light-weight network and is also used in other state-of-the-art lightweight architectures~\cite{sandler2018mobilenet2, howard2017mobilenet, zhang2017shufflenet}. Our proposed design block also uses DWconv. This technique factorizes a standard convolution layer into 3 x 3 depthwise convolution and 1 x 1 pointwise convolution. The depthwise convolution has a very light amount of computation and parameters, since each channel of the input performs a convolution by corresponding to a single depth filter. 1 x 1 pointwise convolution receives this and adjusts the output channel to required amount of network.

MobileNet v2~\cite{sandler2018mobilenet2} proposes a design block which mainly based on inverted residuals, linear bottlenecks and depthwise separable convolutions~\cite{dwconv16chollet}. Also, they proposed the light-weight architecture with shrinking parameters to change the model size, called depth multiplier with a parameter $d$.
ShuffleNet~\cite{zhang2017shufflenet} compress the model by using group convolutions and adopting channel shuffle operation which allows the feature information to interchange actively. They experimentally demonstrated the effectiveness of channel shuffling through a group parameter $g$ along with a light-weight architecture.

We notice that bottleneck block is very efficient method for model compression because the bottleneck block is utilized in MobileNet, ShuffleNet and DenseNet as well. However, in the bottleneck block, the information in the upper layer is transmitted to the bottom layer by adding operation and the block end 1x1 convolution that contract the information. However, what if there is no residual information from the upper layer? For example, in encoder-decoder architecture, the information of a layer in the encoder pass to a layer in the decoder, not to the bottom layer. MobileNet v2~\cite{sandler2018mobilenet2} investigate semantic segmentation~\cite{deeplab32018chen}. Here, they did not consider the layers in the decoder structure in the lightweight view point.

It is necessary to consider that the layer compressed by 1x1 convolution have enough information without skip connection.

\textbf{Justification of Our Design Direction}
Repeating convolution layers has been validated as an effective network design approach. 
Deep VGG~\cite{SimonyanZ14vgg} networks, for instance, are built by adding convolutional layers of similar hyperparameters to achieve very high performance. The number of such convolutional layers ranges from 11 to 19 layers in VGG16: conv-64 layer is repeated twice, conv-128 twice, conv-256 three times, and so on, with max pooling layers inserted between repeated layers.
In this work, we parameterize the repetition as $r$ to simplify the problem of network design and exploration.

On the other hand, GoogLeNet~\cite{szegedy2015going} introduced a block level design that makes network design faster. For instance, GoogLeNet reiterates inception blocks (or modules) nine times. Our approach is  to combine both the layer-wise repetition of VGG~\cite{SimonyanZ14vgg} and the block-based design approach of GoogLeNet~\cite{szegedy2015going}.  That is, our repetition is between pooling and pooling-like layers, similar to VGG network, with a lightweight building block, the RR block, like GoogLeNet.

The main problems with block repetition are the memory size, power consumption, and overfitting issues. 
According to the ``No Free Lunch'' (NFL) theorem~\cite{lunch96davi}, the hypothesis space is limited by the amount of training data. In other words, the limited amount of training data dictates the upper bound for the number of parameters. Therefore, RR blocks should be very lightweight in spite of many iterations. 

From this perspective, we exploit the fact that a 1x1 convolution with expanding layers can reduce not only computational complexity but also the number of parameters. The ability to reduce the number of parameters is particularly crucial in the encoder-decoder architecture to transfer feature information effectively to the decoder. 
Our use of 1x1 convolution layers as the reduction and abstraction mechanism to simplify the data transfer in the encoder-decoder architecture is one of our key contributions. By first compressing features through 1x1 convolution, and passing its output to the decoder, we can vastly decrease the decoder complexity with minimal impact on the network performance.

Recently, many studies have shown that a 1x1 convolution can very effectively reduce the number of computation. The 1x1 convolution also plays an important role in  bottleneck approach~\cite{sutskever2013}, by rearranging the layers to reduce the complexity caused by the kernel window and the channel depth.
GoogLeNet~\cite{szegedy2015going} also adopts 1x1 convolution in its inception module to reduce computational complexity by reducing  weight parameters and the corresponding number of multiplications and accumulations.

\section{Methods}

\begin{figure}[t]
\begin{center}
 \includegraphics[width=0.4\textwidth]{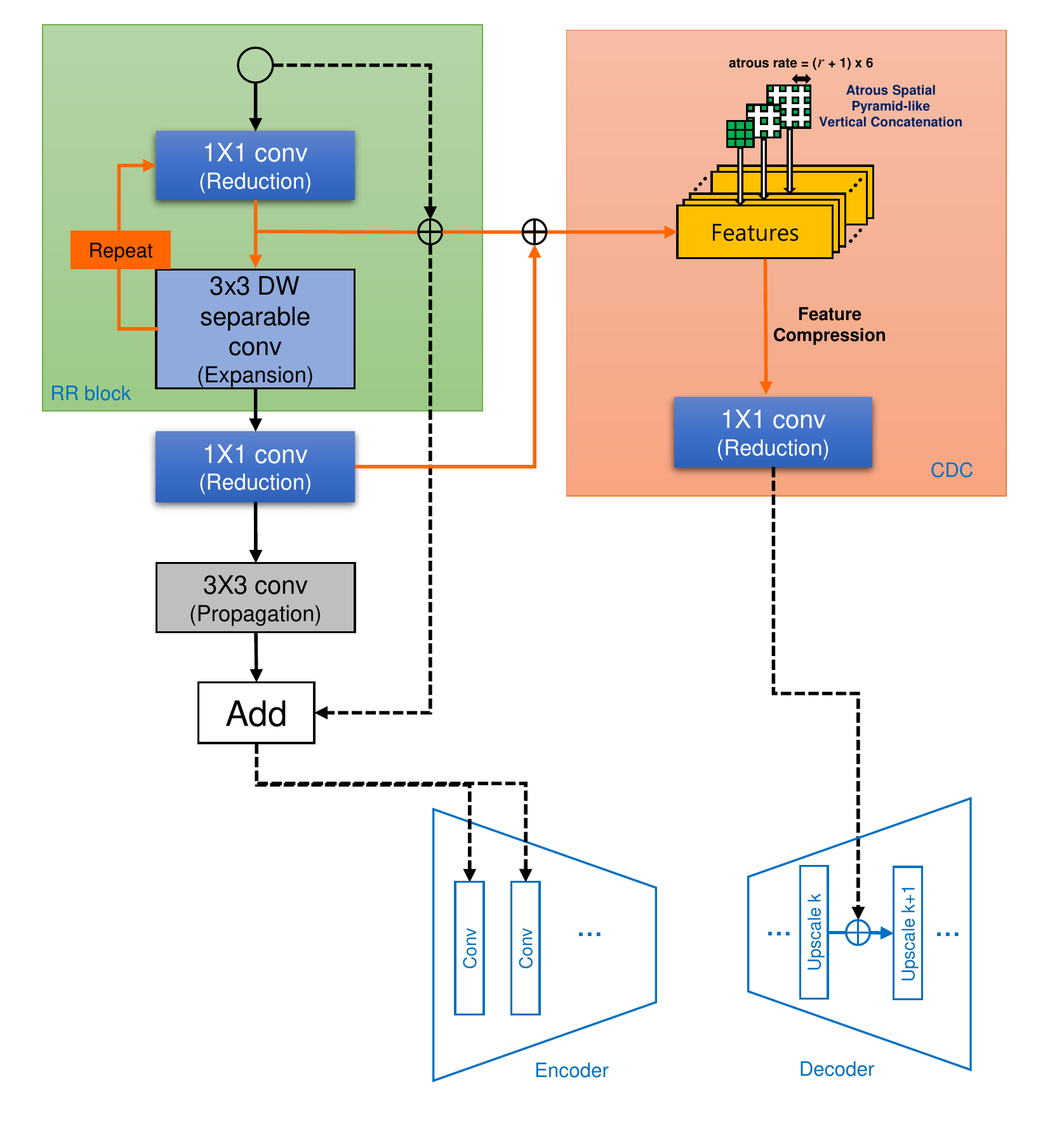}
\end{center}
 \caption{Our proposed RR (Repetition and Reduction) block and CDC.}
\label{fig:rr_block}
\end{figure}

As shown in Figure \ref{fig:rr_block} and as further detailed in this section, the main modules of RRNet can be divided into two sections, namely RR block and CDC module, corresponding to green and red regions, respectively. The RR block located in the encoder performs feature extraction and information propagation by repeat operation, and the  CDC module collects features to be passed to the decoder. 

\subsection{Repetition and Reduction (RR) Block}

In Figure~\ref{fig:rr_block}, the green region highlights the layer repetition component, similar to the basic unit of MobileNet~\cite{sandler2018mobilenet2} and ShuffleNet~\cite{zhang2017shufflenet}. A single unit consists of two layers: a 1x1 convolution layer for reduction and a 3x3 depthwise separable convolution layer for expansion.We adopt an atrous operation in 3x3 depthwise separable convolution layer as shown in Figure~\ref{fig:rr_block}.
 A RR block adopts the bottleneck structure in order to send reduced features to the following CDC module, instead of inverted bottleneck structure in~\cite{sandler2018mobilenet2}.
The bottleneck layer acts as a dimension reduction component, as in principal component analysis(PCA)~\cite{sutskever2013}. To leverage this critical reduction potential, the proposed RRNet intensifies this bottleneck layer by iterating a RR block and stacking each output per repetition. we derive repetition parameter $r$ from the number of repetition of RR blocks. Each iteration can follow two paths. One path is for the next layer in the encoder; the other path leads to the decoder. In the decoder path, each output per iteration is stacked in the CDCs. 
This block is repeated by the parameter $r$. Here, we introduce another two hyperparameters, a reduction parameter $rr$ and a expansion parameter $re$. $rr$ is the number of output channel in stage 1 by pointwise convolution and  $re$ is the number of output channel present after expanding the input through 3x3 depthwise separable convolution (DWconv), which limit the computation or the number of parameters when the channel numbers change. In Figure~\ref{fig:comparison_network}, $rr$ and $re$ are showed in the above the RR block. 
The repetition enriches the model information, resulting in better performance with little changed in Table 1. 

 The RR block has a very efficient structure and can construct CDCs through repetition. A RR block $R(X_i)$ can be denoted by the following formula:
 $R(X_i) = F_{c_{1 \times 1}}(F_{dws-c}(F_{c_{1 \times 1}}(X_i))) \inlineeqnum$.
 CDCs can stack feature maps produced by the RR block. However, the RR block provides a very light-weight unit block for repetition. As shown in the manuscript, a larger number of repetitions does not increase the model size.   
In the Table~\ref{Tb:rrblock}, only the encoders of the two experiments are different: Godard's and RRNet's. 
The RRNet encoder comprised RR blocks. Table~\ref{Tb:rrblock} shows that, by adopting RR blocks in the encoder, the parameters are reduced from 31.6M to 8.8M with a slight degradation in performance.

\begin{table*}[th]
\centering
\caption{Ablation Study of RR block on KITTI 2015}
\small
\resizebox{\textwidth}{!}{
\begin{tabular}{|l|r|r|r|r|r|r|r|r|r|r|r|}
\hline
\textbf{Encoder}&Conn& Decoder& Abs Rel & Sq Rel	& RMSE	& RMSE log	& $\sigma < 1.25$	& $\sigma < 1.25^2$ &  $\sigma < 1.25^3$ & MAdds & Params \\
\hline
\hline
\textbf{Godard}& skip & VGG & 0.0686 & 0.8503 &	4.4060	& 0.1460 & 0.9420 & 0.9770 & 0.9890 & 42.35B & 31.60M \\
\textbf{RRNet} & skip & VGG & 0.0776 & 1.0353 & 4.7370 & 0.1550 & 0.9360 & 0.9750 & 0.9880 & 11.95B & 8.82M\\
\hline
\end{tabular}
}
\label{Tb:rrblock}
\end{table*}

\subsection{Condensed Decoding Connection(CDC)}
The proposed architecture then collects the outputs of the repeated reduction layers as repetition for the condensed decoding connections (CDCs), in the red region in Figure~\ref{fig:rr_block}, and forward them to 1x1 pointwise convolution layer, which we call the reduction layer. This layer plays a very important role, collecting the encoder's feature information and sending it to the decoder. 
In the CDC component, various atrous pyramidal features are stacked separately and forwarded to reduction layer, which handles feature explosion.
The atrous rate is increased at a certain rate according to the number of necessary iterations $r$, thereby maximizing the diversity of features. We used an initial rate as six, which then increased to 12, 18, ... upon repetition. 

The preservation of a limited number of channels with a linear bottleneck is the key to efficiency in MobileNet v2~\cite{sandler2018mobilenet2}. Here, we apply the linear bottleneck to connect not only residuals but also the decoder layers using CDCs. The stacked atrous pyramidal features are compressed by 1x1 reduction layer.  
Then, the reduction layer is delivered to the decoder in a non-activated form that guarantees linearity. Because of repetition, the number of stacked CDC channels is large; for example, 128 channel repeated 6 times is equivalent to 768 channels. We reduce this number by applying the 1x1 reduction layer, resulting the output can be 128 channels, equivalent to $rr$ or $re$ and then connected to the decoder layer. The CDC output is then applied to pointwise 1x1 convolution (called reduction layer) for transmission to the decoder as seen in Figure~\ref{fig:rr_block}. 
Because of a 1x1 convolution, the total output size becomes much smaller than conventional architectures. 
In short, with the smaller unit RR block, enriched CDC features and a 1x1 linear reduction layer, we significantly reduce the number of parameters and energy consumption and obtain better performance.

A CDC is an  enriched version of the skip connection.  
The skip connection can be represented by $ F(x) + x$. 
The ``+" makes the shortest path from the top layer to the bottom layer in the backward propagation process. 
The ``+" are considered as operations to add more edges in the computation graph, which smoothens the loss function and improves its performance.
Furthermore, a greater number of nodes can lead to a more complex computation but the additional edges are relatively smaller 
increase in computation complexity.
Therefore, CDCs add more edges with the same number of nodes in the network. 
Assuming a network with L layers with symmetric skip connections, we denote $F_c$ and $F_d$ as the convolution and deconvolution in each layer with ReLU and $R(X_i)$ is a RR block with input $X_i$. Then, $\Omega(X_i)\!=\! [R(X_{i+1}), R(X_{i+2}), ..., R(X_{i+r})] \inlineeqnum$ where $\Omega$ denotes CDCs. 
We assume information on the convolutional feature map $X_{i}$ is to be passed to the corresponding deconvolutional layer $X_{L-i}$, then $X_{L-i}\!=\!F_d( X_{L-i-1}\!\oplus\!F_{c_{1\times1}}(\Omega(X_i)) \inlineeqnum$ where $\oplus$ denotes the feature map concatenation or a similar operation.
For back-propagation, we consider the $L^{th}$ layer, $X_{L}\!=\!F_d( X_{L-1}\!\oplus\! F_{c_{1\times1}}(\Omega(X_0)) \inlineeqnum$. We compute the derivative of loss $\ell$ with respect to a parameter $\theta$ as follows: $\nabla_{\theta}\ell(X_L)\!=\!\frac{\partial \ell}{\partial X_{L-1}} \frac{\partial X_{L-1}}{\partial \theta} \oplus \frac{\partial \ell}{\partial \Omega(X_0)} \inlineeqnum$. 
Therefore, the gradients corresponding to CDCs carry larger gradients than a only skip connection and it is less likely to approach zero gradients. 

Table~\ref{Tb:cdc} compares the skip connection with our CDCs. Using the same RRNet encoder, we vary the connection type and the corresponding decoder size. In the VGG$\times 1/8$ (``$\times$'' refers depth reduction ratio of each channel) decoder case, the skip connection had 1.2M parameters and the CDCs had 1.16M parameters; however, the performance of the CDCs was better than that of the skip connection. Moreover, CDC$\times 1/16$ had only 0.91M parameters and achieved a higher accuracy than the skip connection with 1.2M parameters.
Owing to the enriched information in CDCs, the network with the fewer parameters can preserve performance.

\begin{table*}[th]
\centering
\caption{Ablation Study of CDC on the KITTI 2015 Dataset}
\small
\resizebox{\textwidth}{!}{
\begin{tabular}{|l|r|r|r|r|r|r|r|r|r|r|r|}
\hline

Encoder& Conn& Decoder& Abs Rel & Sq Rel	& RMSE	& RMSE log	& $\sigma < 1.25$	& $\sigma < 1.25^2$ &  $\sigma < 1.25^3$ & MAdds & Params \\
\hline
\hline
RRNet & skip & VGG$\times 1/8$ & 0.0835 & 1.2131 & 5.0707 &	0.1625&	0.9275&	0.9722&	0.9865& 2.88B & 1.21M\\
\hline
\hline
RRNet & CDC & VGG$\times 1/8$ & 0.0795&	1.0983&	4.9248&	0.1598&	0.9310&0.9740&0.9870& 3.37B & 1.76M\\
RRNet & CDC & VGG$\times 1/16$ &0.0804 & 1.1255	& 4.9085 & 0.1595 &0.9308&0.9740 &0.9868& 3.19B & 1.38M\\
RRNet & CDC$\times 1/8$ & VGG$\times 1/8$ & 0.0808&	1.0959&	4.9565&	0.1598&	0.9295&	0.9728&	0.9870& 2.66B& 1.16M \\
RRNet & CDC$\times 1/32$ & VGG$\times 1/32$ & 0.0843&	1.1002&	4.9880&	0.1640&	0.9220&	0.9710&	0.9860 & 1.82B & 0.78M \\

\hline
\end{tabular}
}
\label{Tb:cdc}
\end{table*}

\subsection{RRNet for Efficient Encoder-Decoder Architecture}
In current encoder-decoder architectures, an increase in the encoder's complexity leads to a rise in the decoder's complexity. By applying the bottleneck approach, the encoder produces a  smaller feature map. To extract information, we expand the next layer using the 3x3 depth-wise separable convolution layer. 

RR blocks can be repeated either vertically, horizontally, or in both directions. Vertical repetition allows our network to go deeper and horizontal repetition enables stacking and concatenation with several convolution layers, emulating the atrous spatial pyramid model~\cite{atrous16chen}.  In other words, by adjusting the a few parameters in the RR block, we can generate various encoder-decoder models with deep and rich structures. 

\begin{figure*}[t]
\begin{center}
 \includegraphics[height=7cm, width=0.99\textwidth]{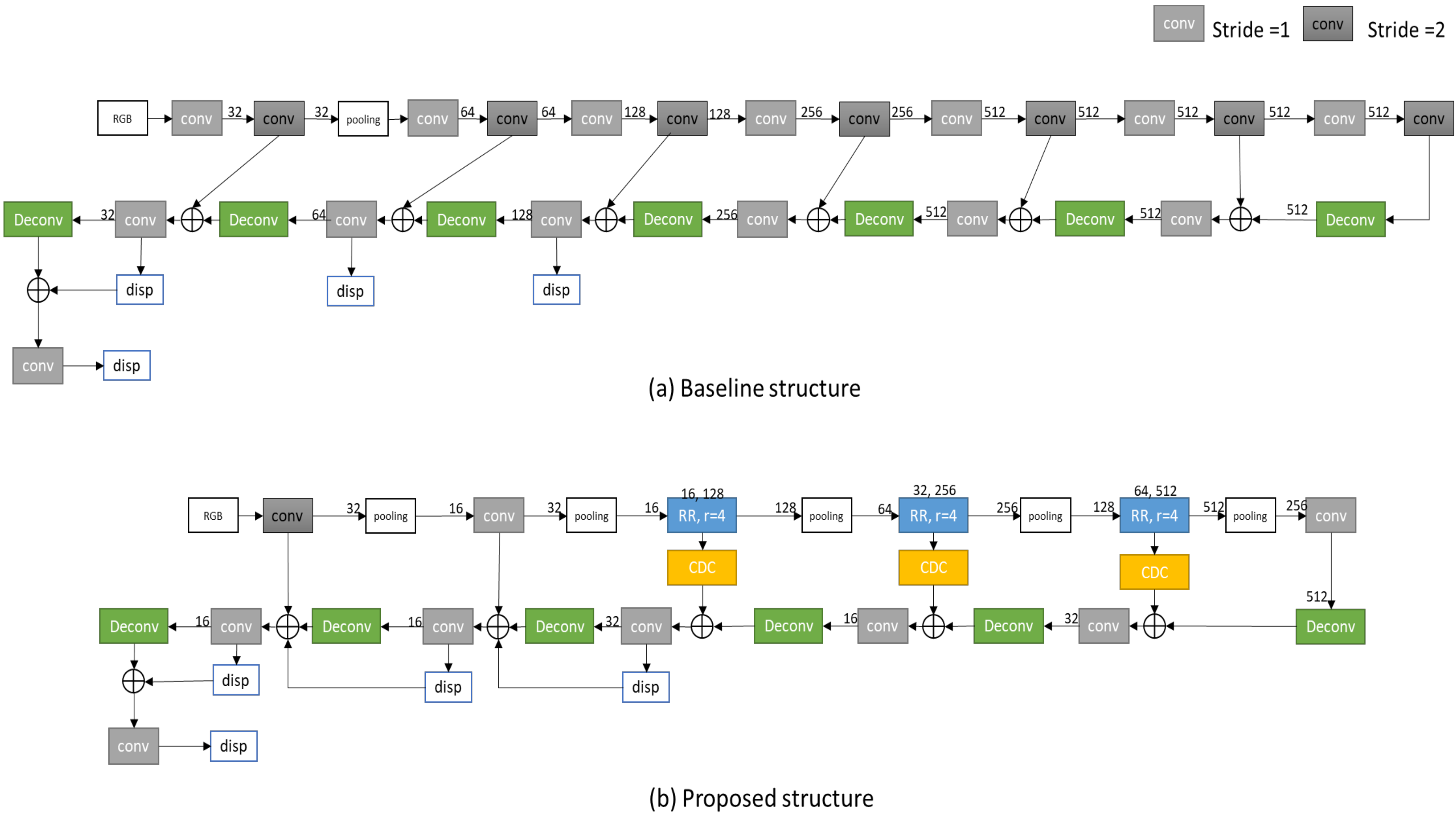}
\end{center}
 \caption{Network Comparisons.  (a) Baseline network\cite{17Godard} (b) Proposed network}
\label{fig:comparison_network}
\end{figure*}

The information flow of RRNet can be divided into an encoding flow and a decoding flow. The encoding flow enhances performance by expanding the amount of information through accumulating convolutional layers with high complexity. Here, the bottleneck layer is arranged in the middle of layers to limit the number of parameters and associated demands on computation resources.

The bottleneck layer allows information to be forwarded appositely into the inflated space even if collected narrowly in a small space. 
The encoder uses the repetition parameter $r$ to create repeatedly bottleneck layers and accumulate them as needed.Thus, the encoder expands the amount of information via layer stacking, in a form that also allows intermediate information gathered in a narrow space to be collected separately.

The decoding flow then receive the gathered information in a expansion space, where the received information simply concatenated to a tensor and then forward through the 1x1 pointwise convolution layer that acts as a reduction layer. This layer acts as an intermediary to enable the encoder's information to be properly extended to the decoder.

\subsubsection{RRNet Decoder}
RRNet Decoder consists of five up-scaling layers. Each layer is based on UNet structure~\cite{unet2015ronneberger}. Up-scaling layer consists of 3 steps: (1) upscale convolution through linear interpolation of feature maps, (2) concatenation of encoder feature information, and (3) 3x3 decoding convolution. The Reconstruction layer from encoder is used here for the concatenation of the second step.

The top three of decoder layers affected by RR blocks of encoder. We created a high-performance model while compressing the decoder very heavily by using CDCs and Reconstruction layers through \textit{repetition} parameter $r$. Model architecture of RRNet Decoder are summarized in Table 2 including 
concatenated CDCs channel and $rcn$ parameter.
Table 2  
describes RRNet decoder in detail.

In MobileNetv2~\cite{sandler2018mobilenet2}, the parameter $d$  is the depth multiplier used as a model shrinking parameter. In ShuffleNet~\cite{zhang2017shufflenet}, the parameter \textit{g}  is a group shuffling parameter for regularization.
We compared the results with varying parameter $d$, $g$ and $r$ , as can be seen in Table 1.

Four $r$ of the RR block (network size: 13M) already achieves the level of performance that is is comparable to MobileNet v2 (size: 59M)

Figure~\ref{fig:comparison_network} shows the detailed baseline architecture and the proposed RRNet structure. The baseline in Figure~\ref{fig:comparison_network} (a) is from Godard's VGG network~\cite{17Godard}. The first row in the Figure~\ref{fig:comparison_network} (a) corresponds to the encoder and the second and third rows are the decoder. For comparisons, SOTA models such as SqueezeNet, MobileNet v2 and ShuffleNet, are used the same decoder as the decoder of the Godard's VGG nework. In Figure~\ref{fig:comparison_network} (b), the decoder of the proposed architecture are much lighter than that of (a) owing to CDCs.

\textbf{Ablation Experiments}
We present ablation experiment of repetition parameter $r$on the KITTI dataset~\cite{Geiger2012CVPR} to justify our design approach. For evaluation, we use the \textit{RMSE} and \textit{Abs Rel} disparity measure, as can be seen in Table 1.

 Repetition parameters are directly related to the density of CDCs and \textit{reconstruction layer} which plays an important role in improving the performance of the whole model. Therefore, evaluation was conducted with various repetition parameters. According to our results, we selected the $r=4$ case for the RRNet considering computation, parameters, and performance. Then all subsequent RRNet experiments were conducted using the $r$=4 case. Results are listed in Table 1. 
 RR Block is so small that the number of parameters do not increase high. 

\begin{table}
\centering
\caption{Comparison of depth estimation results, using monodepth \cite{17Godard} as the baseline encoder-decoder architecture. In RRNet, the \textit{r} value denotes the number of repetition of each RR block. The decoder structure of monodepth \cite{17Godard} is applied to all variations of the SOTA models.}
\resizebox{0.45\textwidth}{!}{
\begin{tabular}{|c|c c c c|}
\hline
Model & RMSE & Abs Rel & MAdds & Params \\
\hline
\hline
monodepth\cite{17Godard} & 4.406 & 0.0686 & 42.36B & 31.60M \\
\hline
SqueezeNetv1.1\cite{iandola2016squeezenet} & 4.638 & 0.0741 & 47.95B & 15.32M \\
\hline
MobileNetv2\cite{sandler2018mobilenet2} & & & &   \\
\textit{d = 1.0} & 4.711 & 0.0732 & 32.46B & 17.60M \\
\textit{d = 0.75} & 4.912 & 0.0786 & 22.27B & 13.03M \\
\textit{d = 0.5} & 4.698 & 0.0736 & 14.51B & 9.36M \\
\textit{d = 0.25} & 4.631 & 0.0729 & 9.18B & 6.59M \\
\textit{d = 0.12} & 4.743 & 0.0759 & 7.60B & 5.55M \\
\textit{d = 0.06} & 4.702 & 0.0755 & 7.21B & 5.11M \\
\hline
ShuffleNet\cite{zhang2017shufflenet}& & & &  \\
\textit{g = 8} & 4.942 & 0.0944 & 10.61B & 17.30M \\
\textit{g = 4} & 4.913 & 0.1050 & 10.77B & 17.74M \\
\textit{g = 3} & 4.867 & 0.1123 & 10.88B & 18.03M \\
\textit{g = 2} & 5.211 & 0.1190 & 11.09B & 18.61M \\
\textit{g = 1} & 5.333 & 0.1205 & 11.72B & 20.36M \\
\hline
RRNet (proposed) & & & &  \\
\textit{$r$ = 1} & 4.931 & 0.0816 & 2.80B & 0.71M \\
\textit{$r$ = 2} & 4.870 & 0.0793 & 2.95B & 0.84M\\
\textit{$r$ = 3} & 4.776 & 0.0771 & 3.11B & 0.97M\\
\textit{r = 4}& 4.539 & 0.0712 & 3.26B & 1.11M  \\
\hline
\end{tabular}
}
\label{Tb:result_d1all}
\end{table}

\section{Experiments}

To evaluate the effectiveness of RRNet, we use unsupervised depth estimation~\cite{17Godard} as our baseline.
Depth estimation provides low-level information, for use by other higher-level applications, and it is frequently executed as a background process.
Therefore, our objective for this evaluation were high performance and minimized runtime and power consumption on mobile devices.

\subsection{Experimental set-up}

\textbf{Dataset:}  For evaluation, the KITTI 2015 dataset~\cite{Geiger2012CVPR} is adopted, which consists of 200 training image pairs and 200 test image pairs. The baseline method~\cite{17Godard} was an unsupervised approach that did not use ground truth depth. KITTI 2015 contains 42,382 rectified stereo pairs from 61 scenes, with 1242x375 pixels. 
We evaluated 200 high qualified disparity images in the training set, covering 28 scenes. The remaining 33 scenes contained 29,000 images for training and 1,159 images for validation. For convenience, we used left and right image together as a single input, and we used 400 images as the test set from KITTI 2015 to evaluate the following variables. 

\textbf{Params (M):} Params is the total number of trainable parameters and we obtained this value using the profiler function provided by TensorFlow\cite{abadi2016tensorflow}.

\textbf{MAdds (B):} The number of multiplication and addition operations executed during the inference task, referring to the evaluation of all 400 images from KITTI 2015. We also got this value through the profiler function of the TensorFlow.

\textbf{Memory (MB):} The average amount of system memory occupied by the application during inference task. The reason why the target is system memory is because TX2 shares system memory instead of local device memory.

\textbf{Runtime (sec):} The total execution time consumed by the application during the inference task.

\textbf{Power (W):} The average number of watts consumed by the application during the inference task.

\textbf{Energy (J):} The total energy consumption by the application expressed in joules during inference task.

{\textbf{Evaluation Metric}} We adopted the same metrics used in~\cite{17Godard}. Given a ground truth depth $D$ = {d} and predicted depth $\hat{D}$ ={$\hat{d}$}, we evaluated the results using the following metrics: 

(1) RMSE (Linear Root Mean Square Error):

$\sqrt{\frac{1}{|\hat{D}|}  \sum_{\hat{d} \in \hat{D} } \Arrowvert d - \hat{d} \Arrowvert ^2}$ , 

(2) RMSE log (Log scale Invariant RMSE): 

$\frac{1}{|\hat{D}|}  \sum_{\hat{d} \in \hat{D} } (log(d) -log (\hat{d}))$,

(3) Abs Rel (Absolute Relative Error): 

$\frac{1}{|\hat{D}|}  \sum_{\hat{d} \in \hat{D} } |d - \hat{d} |/d $ ,
 
(4) Sq Rel (Square Relative Error): 

$\frac{1}{|\hat{D}|}  \sum_{\hat{d} \in \hat{D} } (d - \hat{d} )^2/d $
 
(5) $ \delta_t$ (Inlier Ratio): 

$\hat{d} \in \hat{D}$ s.t. $max(\frac{d}{\hat{d}}, \frac{\hat{d}}{d} ) < t $ where $t \in {1.25, 1.25^2, 1.25^3}$ 

\textbf{Training details:} RRNet was trained using TensorFlow~\cite{abadi2016tensorflow} with CUDA 8.0 and cuDNN 7.0 back ends. We assessed RRNet's performance with respect to the results reported in   Godard~\cite{17Godard} and with PyDNet~\cite{Poggi18}.  Godard's model was used as the baseline for evaluations of both PyDNet~\cite{Poggi18} and RRNet. For a fair comparison, we trained our network with the same protocol as used in ~\cite{17Godard,Poggi18}: batches of eight images resized to 256x512x3, ececuting 300 epochs on 29,000 images. Our loss function and hyperparameters are also preserved in~\cite{17Godard, Poggi18}.

In~\cite{17Godard}, there were two modes for training, namely \textit{mono} and \textit{stereo}. These modes are selected according to the number of inputs:  the \textit{mono} mode for a single input and the \textit{stereo} mode for image inputs by concatenation. 
In general, stereo approaches outperform monocular approaches via information enrichment. However, stereo can significantly increase memory usage, runtime and energy consumption. To show the effectiveness of RRNet, we adopt the \textit{stereo} mode for evaluation. 

Table 2 compares our RRNet with SqueezeNet~\cite{iandola2016squeezenet}, ShuffleNet~\cite{zhang2017shufflenet}, MobileNet v2~\cite{sandler2018mobilenet2} and PyDNet~\cite{Poggi18}, which are representative lightweight architectures.
For the best case referenced in Table 1, the proposed method decrease the amount of computation is decreased by 4.16 $\sim$ 13.9 times and the number of parameters by 4.53 $\sim$ 16.6 times.

These experiments were performed in a resource-constrained mobile edge device, NVIDIA Jetson TX2 (256 CUDA cores). 
To evaluate hardware-level statistics such as power consumption and memory, we used \textit{tegrastats}, which is a built-in executable file in TX2 to obtain all memory statistics, and we implemented an energy measurement script using the system variables associated with the internal power sensors accessible in the TX2 operating system.

For evaluation on other datasets, the Cityscapes dataset is adopted and its results are shown in Table~\ref{Tb:city_result}.
\vspace{-2mm}
\begin{table*}[th]
\centering
\caption{Other Dataset Results: Cityscapes}
\small
\resizebox{\textwidth}{!}{
\begin{tabular}{|l|c|r|r|r|r|r|r|r|r|r|}
\hline
\textbf{Model}&Dataset & Abs Rel & Sq Rel	& RMSE	& RMSE log	& $\sigma < 1.25$	& $\sigma < 1.25^2$ &  $\sigma < 1.25^3$ & MAdds & Params \\
\hline
\hline
\textbf{Godard} & CS + K & 0.0721&	0.9892&	4.5420&	0.1500&	0.9430&	0.9760&	0.9870 & 42.36B & 31.60M\\
\textbf{PyDNet} & CS + K & 0.0897 & 1.1791 & 5.0560	& 0.1710&0.9210 &0.9700 &0.9850 & 9.84B & 1.97M\\
\textbf{RRNet} & CS + K &  0.0781&	1.0706&	4.8598&	0.1573&	0.9343&	0.9745&	0.9870 & 3.26B & 1.11M\\
\hline
\end{tabular}
}
\label{Tb:city_result}
\end{table*}

\begin{figure}[t]
\begin{center}
 \includegraphics[width=0.47\textwidth]{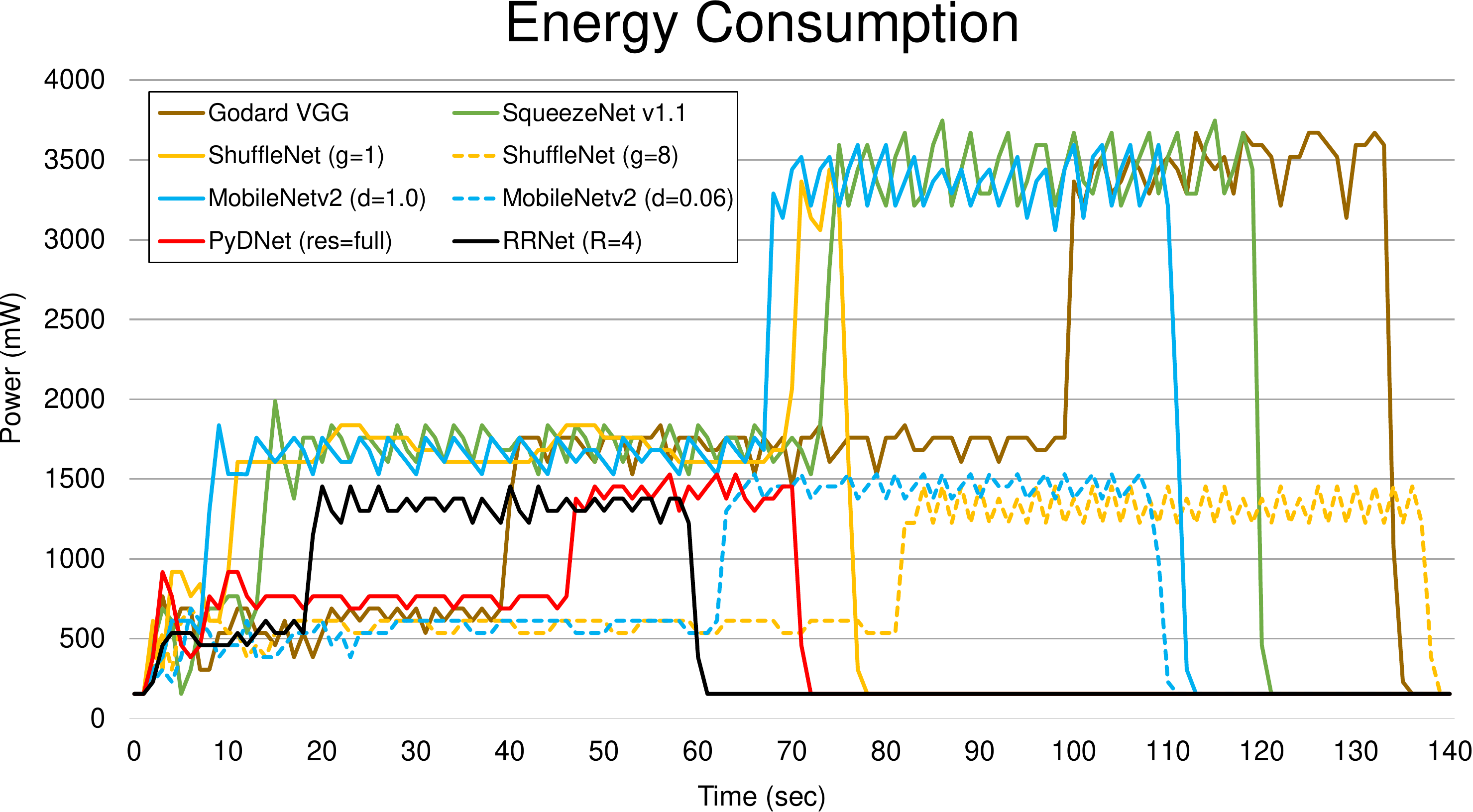}
\end{center}
 \caption{Energy Consumption Results on NVIDIA TX2.}
\label{fig:result_tx2_power}
\end{figure}

\begin{figure}[t]
\begin{center}
 \includegraphics[width=0.47\textwidth]{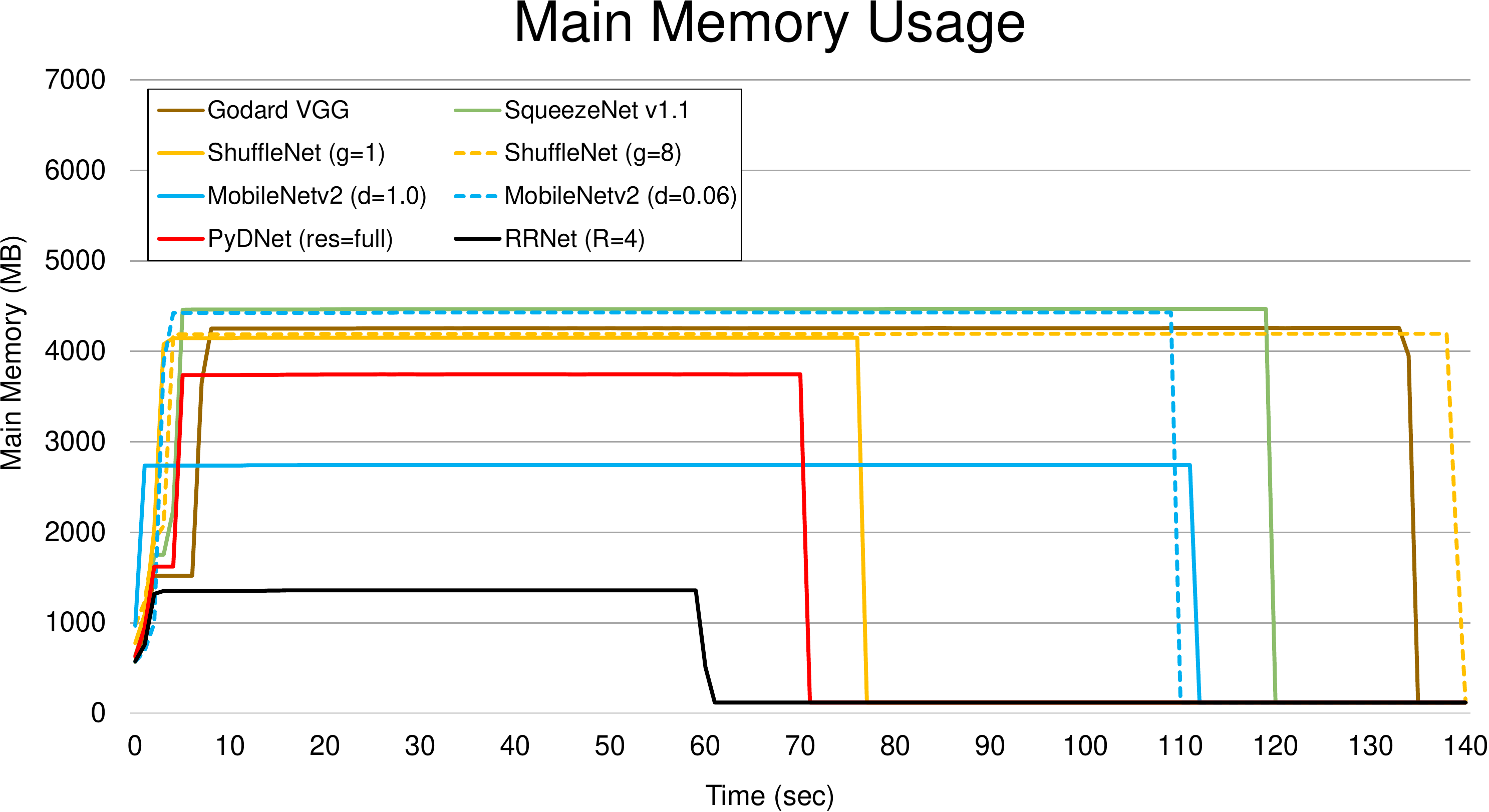}
\end{center}
 \caption{Main Memory Usage Results on NVIDIA TX2.}
\label{fig:result_tx2_memory}
\end{figure}

\begin{table*}
\begin{center}
\caption{Results on KITTI 2015 on a NVIDIA TX2. We compared models from Table 3 that has the highest or lowest complexity in terms of computation, parameters or accuracy. Memory refers to the average consumed main memory consumed by model execution. Runtime refers to execution time for 400 images from KITTI 2015.}
\small
\resizebox{\textwidth}{!}{
\begin{tabular}{|l|r||r|r|r|r|r|r|r|}
\hline
Application & Baseline \cite{17Godard} & SqueezeNet \cite{iandola2016squeezenet} & ShuffleNet \cite{zhang2017shufflenet} & ShuffleNet \cite{zhang2017shufflenet} & MobileNetv2 \cite{sandler2018mobilenet2} & MobileNetv2 \cite{sandler2018mobilenet2}& PyDNet~\cite{Poggi18} & Ours (RRNet) \\
 & & \textit{v1.1} & \textit{g = 8} & \textit{g = 1} & \textit{d = 1.0} & \textit{d = 0.06} & \textit{res = full} & \textit{r = 4}\\
\hline\hline
Params (M) & 31.60 & 15.32 & 17.30  & 20.36 & 17.60 & 5.11 & 1.97 & \textbf{1.11}\\
MAdds (B) & 42.36 & 47.95 & 10.61 & 11.72 & 32.46 & 7.21 & 9.84 & \textbf{3.26}\\
Memory (MB) & 3926 & 4238 & 3890 & 4072 & 2658 & 4211 & 3431 & \textbf{1281}\\
Runtime (s) & 137 & 122 & 79 & 140 & 114 & 112 & 73 & \textbf{62}\\
Energy (J) & 246.03 & 268.89 & 126.41 & 120.86 & 249.84 & 101.49 & 67.80 & \textbf{64.07} \\
\hline \hline
Abs Rel & 0.0686 &	0.0741	&0.0944	&0.1205&	0.0732	&0.0755&	0.0935 &\textbf{ 0.0712}  \\
Sq Rel &  0.8503	&1.1088	&1.1071&	1.1427	&1.1506&	1.0113	&1.2222 &\textbf{0.8939}	 \\
RMSE & 4.406&	4.638&	4.942&	5.333&	4.711&	4.702&	5.118& \textbf{ 4.539}	\\
RMSE log & 0.146&\textbf{	0.151}&	0.169&	0.206&	0.155&	0.152&	0.175 &\textbf{0.151}	\\
$\delta < 1.25 $  & 0.942&	\textbf{0.943}&	0.92&	0.869&	\textbf{0.943}&	0.939&	0.917& 0.94	 \\
$\delta < 1.25^2 $ &  0.977&	0.977&	0.972&	0.954&	0.976&\textbf{	0.977}&	0.968& 0.976	 \\
$\delta < 1.25^3 $ &  0.989&\textbf{	0.988}&	0.986&	0.978&	0.987&	\textbf{0.988}&	0.984 &\textbf{ 0.988}	\\

\hline
\end{tabular}
}
\end{center}
\label{Tb:result_all_tx2}
\end{table*}

\begin{figure*}[h]
\begin{center}
 \includegraphics[width=\textwidth]{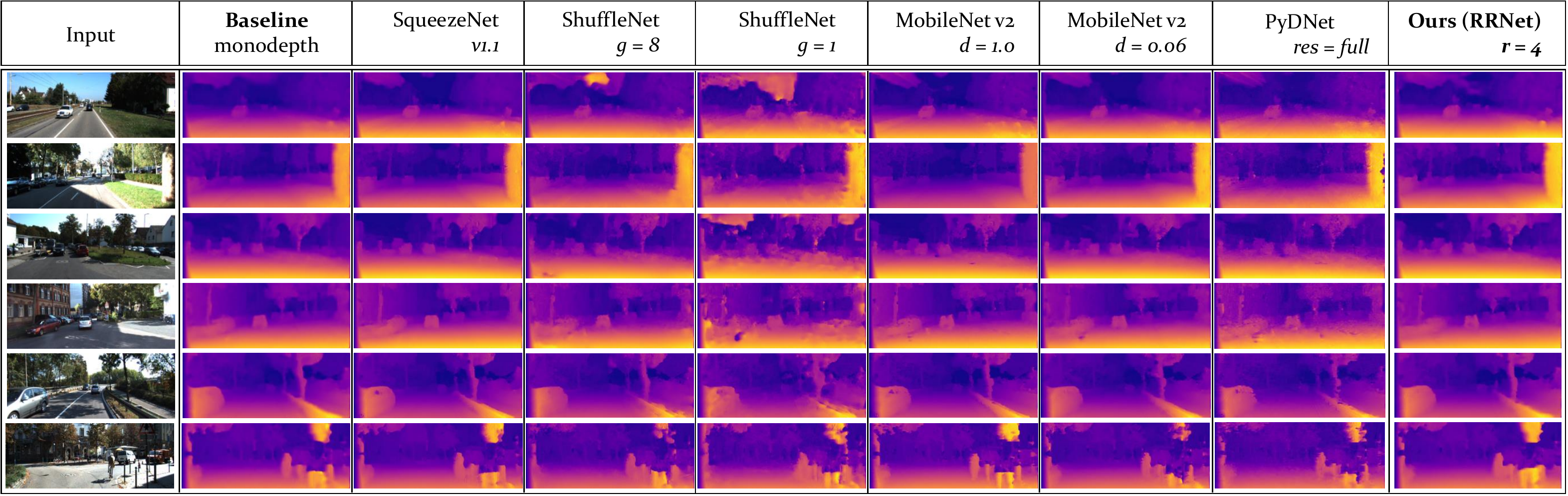}
\end{center}
 \caption{TX2 results on KITTI 2015.}
\label{fig:tx2_results}
\end{figure*}

\subsection{Evaluation RRNet on NVIDIA TX2}

For resource-constrained mobile applications, the model size, runtime, and the energy consumption are key considerations. We evaluated these on NVIDIA TX2 Development Kit~\cite{JetsonTX2} by using depth estimation based on ARM-A57 CPU with 8GB of main memory running on a Linux OS, namely Ubuntu 16.04.

TX2 is a mobile GPU core based on the Pascal architecture, which has 0.75 TFLOPS, and 7 W of thermal design power. We rebuilt common workstation environment, using TensorFlow r1.4~\cite{abadi2016tensorflow}, CUDA 8.0, and cuDNN 7.0, the only difference being that TensorFlow was custom-built to be run on the ARM architecture.

The results of running RRNet and other architectures on TX2 are summarized in Table 2. 
RRNet had a total energy consumption of 64.07J, which was the lowest consumption among state-of-the-art architectures and it required approximately 3.84 times less power and 2.21 times faster than the baseline~\cite{17Godard}. RRNet also outperformed in terms of memory usage and the amount of computation, as shown in Figure~\ref{fig:result_tx2_memory}.

We emphasize our energy consumption because the energy is affected by the runtime factor.  In the real world, if the application is executed for an extended length of time, the difference of energy consumption and power increase and RRNet had the shortest runtime as seen in Figure ~\ref{fig:result_tx2_memory}. 

Therefore, we expect our method to have even greater benefit in such cases.
We also analyzed the depth estimation result qualitatively, as shown in Table~\ref{Tb:result_all_tx2}.
RRNet with $r=4$ showed the best performance in terms of most evaluation metrics listed above except for $\delta<1.25$ with 0.003 difference and $\delta<1.25^2$ with 0.001 degrade. Moreover, the last row in Figure 4 
suggests that the close-range traffic sign was well-recognized using the proposed method, compatible with baseline case~\cite{17Godard} which is a highly complex model.   

\section{Conclusion}

In this paper we proposed a RR block as the building blocks for a lightweight encoder-decoder networks. Our RR block is very efficient in terms of computational complexity and parameter size,  facilitating improved design for   encoder-decoder architectures. 
RRNet makes not only the encoder but also the decoder very lightweight. The proposed architecture is small enough to apply to mobile devices and differs from the previous encoder-decoder enhancement approach~\cite{sandler2018mobilenet2,DBLP:rtseg2018siam}. 

RR Blocks can increase the effectiveness of the network designs, particularly for an encoder-decoder architecture. RR blocks can be stacked in feature maps with different scales and can also reduce feature maps by using pointwise convolutions with the CDCs, as shown in Figure 1.
The RR block can be repeated without burdening the memory issue. The RR block repetition direction can be either horizontal, vertical, or both. Through concatenating, atrous spatial pyramid type architecture~\cite{atrous16chen} can also be supported by the proposed architecture, which can effectively compress feature information of the encoder in an RR block and reduce it in the decoder. We show such a scheme is pivotal in making lightweight and high-performance network designs.

We also introduced our backbone network RRNet, which is an encoder-decoder model that is very lightweight thanks to our RR Block.
On a commercial mobile GPU, RRNet outperforms previous state-of-the-art models speeding the runtime by approximately 2.21 times, the energy savings of up to 3.84 times and memory savings of up to 3.06 times , with optimal performance. 

We plan to apply the proposed RR block and RRNet to other applications such as semantic segmentation and object detection to evaluate its generalization capability.

{\small
\bibliographystyle{ieee}

}

\end{document}